\documentclass{article}
\usepackage{ijcai24}

\usepackage{times}
\usepackage{soul}
\usepackage{url}
\usepackage[hidelinks]{hyperref}
\usepackage[utf8]{inputenc}
\usepackage[small]{caption}
\usepackage{graphicx}
\usepackage{amsmath}
\usepackage{amsthm}
\usepackage{booktabs}
\usepackage{multirow}
\usepackage{algorithm}
\usepackage{algorithmic}
\usepackage{enumitem}
\usepackage{textcomp}
\usepackage{colortbl} 
\usepackage{xcolor}
\usepackage[switch]{lineno}

\title{Few-shot Learning on Heterogeneous Graphs: Challenges, Progress, and Prospects}

\author{
Pengfei Ding
\and
Yan Wang\and
Guanfeng Liu\\
\affiliations
Macquarie University\\
\emails
\{pengfei.ding, yan.wang, guanfeng.liu\}@mq.edu.au
}

\begin{document}

\maketitle

\begin{abstract}
Few-shot learning on heterogeneous graphs (FL HG) is attracting more attention from both academia and industry because prevailing studies on heterogeneous graphs often suffer from label sparsity. FLHG aims to tackle the performance degradation in the face of limited annotated data and there have been numerous recent studies proposing various methods and applications. In this paper, we provide a comprehensive review of existing FLHG methods, covering challenges, research progress, and future prospects. Specifically, we first formalize FLHG and categorize its methods into three types: single-heterogeneity FLHG, dual-heterogeneity FLHG, and multi-heterogeneity FLHG. Then, we analyze the research progress within each category, highlighting the most recent and representative developments. Finally, we identify and discuss promising directions for future research in FLHG. To the best of our knowledge, this paper is the first systematic and comprehensive review of FLHG.
\end{abstract}

\section{Introduction}
Heterogeneous graphs (HGs), consisting of diverse types of nodes and diverse types of edges, have been widely used to model complex real-world systems, such as social networks \cite{dong2012link}, biological networks \cite{ma2023single}, and e-commerce networks \cite{liu2022survey}. A fundamental and crucial research area within HGs is heterogeneous graph representation learning (HGRL) \cite{wang2022survey}, which aims to utilize deep neural networks (e.g., heterogeneous graph neural networks \cite{bing2023heterogeneous}) to automatically generate low-dimensional yet informative representations of nodes, edges, and subgraphs within HGs \cite{dong2020heterogeneous}. Such representations have been proven to be highly effective for a range of graph mining tasks \cite{yang2023simple}.

\begin{figure}
\centering
\scalebox{0.303}{\includegraphics{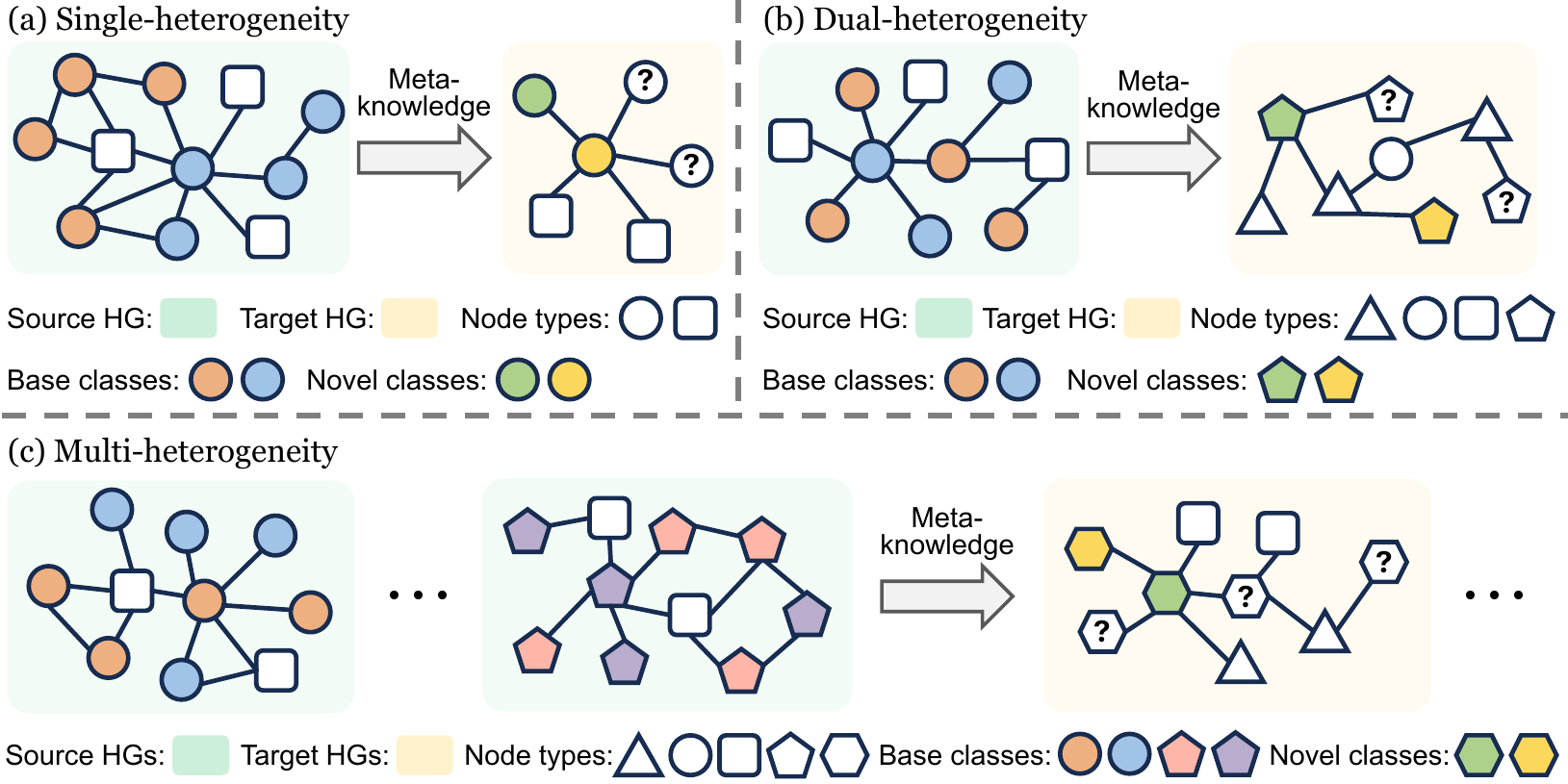}}
\caption{An illustrative example to show different FLHG scenarios.}
\label{eg1}
\end{figure}

\noindent\textbf{Motivation: Why few-shot learning on HGs?}
Existing HGRL models typically require a substantial amount of labeled data for effective training \cite{zhang2022few}. However, label sparsity is a common issue in HGRL because labeling nodes, edges, or subgraphs in HGs demands considerable expertise and resources \cite{yoon2022zeroshot}. To mitigate the performance degradation of HGRL methods in the face of label sparsity, few-shot learning on heterogeneous graphs (FLHG) has been developed to reduce the dependence on extensive labeled data \cite{xie2020heterogeneous}. As illustrated in Figure \ref{eg1}, a FLHG scenario involves source and target HGs. Each HG consists of various types of nodes and some nodes are labeled with different classes. For instance, citation networks typically include node types like “paper” and “author”, with some “paper” nodes potentially classified as \textit{biology} or \textit{physics} based on their research content. The source and target HGs contain nodes from different classes, i.e., \textit{base classes} and \textit{novel classes}, respectively. FLHG aims to extract generalized knowledge (a.k.a., \textit{meta-knowledge}) from base classes that have rich-labeled nodes, and then utilize this knowledge to facilitate learning novel classes with few-labeled nodes. Recently, FLHG has been successfully applied to various applications, such as few-shot node classification \cite{ding2023cross}, few-shot relation prediction \cite{chen2022meta}, and few-shot molecule classification \cite{guo2021few}.

\noindent\textbf{Taxonomy of FLHG: A heterogeneity perspective.} 
\textit{Heterogeneity} in an HG refers to the diversity and uniqueness of node types and edge types within the graph. Two HGs have the same heterogeneity if they share identical node and edge types. In FLHG, differences in heterogeneity between source and target HGs introduce three challenging research scenarios: \textit{single-heterogeneity}, \textit{dual-heterogeneity}, and \textit{multi-heterogeneity}. Conventional FLHG methods typically focus on the single-heterogeneity scenario, where source and target HGs have exactly the same node and edge types (as shown in Figure \ref{eg1}(a)). In this scenario, meta-knowledge about the shared node/edge types can be extracted from the source HG and transferred to the target HG to learn novel classes.

However, in real-world applications, source and target HGs may not always exhibit the same heterogeneity. It is common for these HGs to include different types of nodes and edges. In the dual-heterogeneity scenario, the source HG displays one type of heterogeneity, while the target HG exhibits another (as shown in Figure \ref{eg1}(b)). Specifically, these two HGs either have no overlapping node/edge types, or have only a subset of node/edge types in common. The multi-heterogeneity scenario involves a set of source HGs that exhibit at least two types of heterogeneity, each of which differs from the heterogeneity of the target HG (as shown in Figure \ref{eg1}(c)). In both scenarios, understanding the relationships between different heterogeneities is crucial for extracting relevant meta-knowledge to support learning novel classes. To this end, several FLHG methods \cite{ding2023cross,ding2024few} have been recently developed to adopt different frameworks for addressing problems in these two challenging scenarios.


\noindent\textbf{Motivation of This Survey:} 
In contrast to the numerous recent advancements in FLHG \cite{hao2023heterogeneous,ding2024few}, which include a range of methods and applications across the three scenarios discussed above, there is a lack of a comprehensive review that summarizes these advancements and applications within FLHG. Although several recent reviews have summarized studies related to few-shot learning with graph-structured data \cite{mandal2022metalearning,zhang2022few}, they mainly focus on homogeneous graphs that consist of only a single type of nodes and a single type of edges, overlooking the exploration of more complex FLHG problems. This gap in the literature motivates us to systematically analyze the challenges within FLHG and summarize the corresponding progress in this area.

\noindent\textbf{Our Contributions:} The notable contributions of our paper are summarised as follows: 

\begin{itemize}[leftmargin=*]
\item \textbf{New Taxonomy:} We systematically categorize FLHG methods from a heterogeneity perspective and analyze the challenges of each category, which can help researchers better understand the research landscape.

\item \textbf{Comprehensive Review:} We provide a thorough review of current research progress in FLHG. Specifically, our analysis focuses on the contributions of existing methodologies, as well as the similarities and differences among them.

\item \textbf{Future Prospects:} We discuss some future research directions of FLHG, offering insights that can guide the development of this promising area.
\end{itemize}

\section{FLHG Preliminaries and Challenges}\label{sec2}
FLHG, as a new and promising research area within the AI community, non-trivially combines the strengths of HG representation learning and few-shot learning. In this section, we first provide a brief background on HG representation learning and few-shot learning. Then, we present a detailed formulation of the FLHG problem across three distinct scenarios: single-heterogeneity, dual-heterogeneity, and multi-heterogeneity. We introduce these particular FLHG scenarios and discuss the unique challenges presented in each scenario.

\subsection{HG Representation Learning}
Heterogeneous graph representation learning (HGRL) aims to leverage rich semantics within HGs to generate low-dimensional embeddings of nodes for a variety of graph mining tasks. Recently, most HGRL methods are based on heterogeneous graph neural networks (HGNNs) \cite{bing2023heterogeneous}. These methods extend graph neural networks to handle heterogeneous information within HGs and demonstrate remarkable performance in learning representations on HGs. Specifically, HGRL methods can be categorized into two main types: Meta-path based methods and meta-path free methods. (1) Meta-path based methods \cite{dong2017metapath2vec,wang2019heterogeneous,fu2020magnn} utilize meta-paths \cite{sun2011pathsim} to aggregate information from type-specific neighborhoods. These methods can capture higher-order semantic information specified by the selected meta-paths. However, selecting appropriate meta-paths often requires expert knowledge and significantly influences the model's effectiveness. (2) Meta-path free methods \cite{hu2020heterogeneous,zhu2019relation,hong2020attention} eliminate the dependence on predefined meta-paths. These methods employ a message-passing mechanism directly on the HG with node/edge type-aware modules, allowing the model to simultaneously capture structural and semantic information without the requirement of expert-driven meta-path selection. 

\subsection{Few-shot Learning}
Few-shot learning (FSL) aims to extract generalized knowledge (a.k.a., meta-knowledge) from existing tasks and then apply this knowledge to new tasks with few-labeled data. Existing FSL methods can be mainly divided into two categories: (1) Gradient-based methods \cite{andrychowicz2016learning,finn2017model,rusu2018meta} that focus on optimizing the initial parameters of models. These parameters can be quickly adapted to new few-shot tasks through several gradient updates. (2) Metric-based methods \cite{koch2015siamese,vinyals2016matching,snell2017prototypical} that leverage similarity metrics among samples in existing tasks to classify novel classes in few-shot tasks.

FSL employs an episodic training process where each episode focuses on a specific task. Specifically, each task includes a support set and a query set. The objective of a task is to predict the classes of query set samples using the support set. Typically, after completing sufficient training iterations over all tasks for the base classes (i.e., meta-training), the model developed during this phase is expected to solve each few-shot task on novel classes by classifying samples in the query set, utilizing a limited number of labeled examples per class from the support set (i.e., meta-testing).

\subsection{FLHG Problems}\label{Problems}
Different FLHG problems share a common setting of few-shot learning. Specifically, let $\mathcal{C}$ denote the entire classes set of the whole dataset, which can be divided into two groups: the base class set $\mathcal{C}_\emph{base}$ used for meta-training, and the novel class set $\mathcal{C}_\emph{novel}$ used for meta-testing, where $\mathcal{C}$ = $\mathcal{C}_\emph{base}\cup\mathcal{C}_\emph{novel}$ and $\mathcal{C}_\emph{base}\cap\mathcal{C}_\emph{novel}$ = $ \emptyset$. Generally, the number of labeled data for $\mathcal{C}_\emph{base}$ is abundant while scarce for $\mathcal{C}_\emph{novel}$. 

In this section, we introduce three typical FLHG scenarios and discuss the challenge(s) within each scenario. We define $\mathcal{C}$ as the node class set and use the classical node classification problem as an example to illustrate these scenarios. Note that $\mathcal{C}$ can also represent the class set for other data types in various HG problems, such as the edge class set for the relation prediction problem \cite{xie2020heterogeneous}, and the graph class set for the graph classification problem \cite{guo2021few}.

\noindent\textbf{Definition 1. {Single-heterogeneity FLHG:}} {Given a source HG $\textit{G}_\textit{src}$ containing labeled nodes from $\mathcal{C}_\textit{base}$, and a target HG $\textit{G}_\textit{tgt}$ with labeled nodes from $\mathcal{C}_\textit{novel}$. Both $\textit{G}_\textit{src}$ and $\textit{G}_\textit{tgt}$ exhibit the same heterogeneity, i.e., they share identical sets of node types $\mathcal{A}_\textit{src}=\mathcal{A}_\textit{tgt}$ and edge types $\mathcal{R}_\textit{src}=\mathcal{R}_\textit{tgt}$, where $\mathcal{A}$ and $\mathcal{R}$ represent the sets of node and edge types, respectively. FLHG in this single-heterogeneity scenario aims to develop a machine learning model such that after training on labeled nodes in $\mathcal{C}_\textit{base}$, the model can accurately predict labels for nodes within the query set that belong to $\mathcal{C}_{\textit{novel}}$, using only a small number of labeled nodes from the support set.

Single-heterogeneity is a common scenario in the FLHG area, with the majority of existing methods targeting this scenario. In some special cases, $\textit{G}_\textit{src}$ and $\textit{G}_\textit{tgt}$ may refer to the same HG \cite{zhuang2021hinfshot}. Based on several recent and representative studies \cite{zhang2022fewcross,hao2023heterogeneous}, we identify two significant challenges associated with this scenario as follows:

\noindent\textbf{Challenge 1. Extracting appropriate meta-knowledge for specific tasks (CH1).} 
In this scenario, the source HG and the target HG share numerous commonalities, including similar node features, edge features, and semantics, along with the same types of nodes and edges. Mining meta-knowledge from these shared elements is beneficial for learning $\mathcal{C}_\textit{novel}$ with few-labeled data. However, the contribution of these heterogeneous elements often varies across different tasks, and extracting generalized knowledge from all these commonalities can increase the complexity of the model. Therefore, how to identify the most relevant heterogeneous elements from these commonalities and extract the appropriate meta-knowledge for specific tasks present a fundamental and significant challenge.

\noindent\textbf{Challenge 2. Supporting few-shot tasks with auxiliary tasks from unlabeled data (CH2).}
In HGs, apart from few-labeled data, unlabeled nodes, edges, and subgraphs also contain rich information, such as graph structure connectivity, node clustering, and meta-path connections \cite{wang2021self}. Such information can be leveraged to regularize model parameters or enhance representation learning in few-shot scenarios. Therefore, how to design auxiliary self-supervised/unsupervised learning tasks from unlabeled data and effectively combine these tasks with few-shot learning presents a necessary and compelling challenge.

\noindent\textbf{Definition 2. Dual-heterogeneity FLHG:} Given two different HGs $\textit{G}_\textit{src}$ and $\textit{G}_\textit{tgt}$ that include labeled nodes from $\mathcal{C}_\textit{base}$ and $\mathcal{C}_\textit{novel}$, respectively, $\textit{G}_\textit{src}$ and $\textit{G}_\textit{tgt}$ have different heterogeneities, i.e., $\mathcal{A}_\textit{src}\neq\mathcal{A}_\textit{tgt}$ and $\mathcal{R}_\textit{src}\neq\mathcal{R}_\textit{tgt}$. FLHG in this dual-heterogeneity scenario aims to develop a machine learning model such that after training on labeled nodes in $\textit{G}_\textit{src}$, the model can accurately predict labels for nodes (query set) in $\mathcal{C}_\textit{novel}$ with only limited labeled nodes (support set).

Different from the single-heterogeneity FLHG, the dual-heterogeneity FLHG aims to transfer meta-knowledge across source and target HGs that consist of different types of nodes and edges. From our recent research \cite{ding2023cross,ding2024few}, we have found that dual-heterogeneity FLHG approaches face two primary challenges as follows:

\noindent\textbf{Challenge 3. Mitigating distribution shifts between source and target HGs (CH3).} The heterogeneity differences between the source HG and the target HG inevitably lead to distribution shifts between the two graphs. These shifts are multifaceted, involving variations in node/edge features, changes in graph sizes, and discrepancies in node/edge types. Such distribution disparities can render the meta-knowledge extracted from the source HG less effective when applied to the target HG. Therefore, it is crucial to determine how to mitigate the impact of these distribution shifts and discover meta-knowledge that remains invariant despite these shifts.

\noindent\textbf{Challenge 4. Identifying common semantics between heterogeneities (CH4).} In the dual-heterogeneity scenario, the source HG and the target HG often lack apparent similarities, such as common node and edge types. This absence of evident commonalities calls for the exploration of more complex semantic information to discover the connections and commonalities between the two distinct heterogeneities. Since mining underlying semantics from HGs usually requires specialized domain knowledge, a key challenge lies in how to automatically mine the semantic information of each heterogeneity and reveal shared semantics between them.

\noindent\textbf{Definition 3. {Multi-heterogeneity FLHG:}} 
Given a target HG $\textit{G}_\textit{tgt}$ with its novel class set $\mathcal{C}_\textit{novel}$, there exists a set of source HGs $\mathbf{G_\textit{src}}$ = $\{\textit{G}_1, \ldots, \textit{G}_m\} (m\geq 2)$. Each $\textit{G}_i\in \mathbf{G_\textit{src}}$ has its own node class set $\mathcal{C}_i$, node type set $\mathcal{A}_i$, and edge type set $\mathcal{R}_i$. The base class set across these source HGs is defined as $\mathcal{C}_\textit{base}$ = $\{c|c\in \mathcal{C}_i, \textit{G}_i \in \mathbf{G_\textit{src}}\}$. In addition, each source HG has distinct graph structure and heterogeneity from the target HG, i.e., $\textit{G}_\textit{tgt} \notin \mathbf{G_\textit{src}}$, $\mathcal{A}_i \neq \mathcal{A}_{tgt}$ and $\mathcal{R}_i \neq \mathcal{R}_{tgt}$. In this multi-heterogeneity scenario, FLHG aims to develop a machine learning model such that after training on labeled nodes in $\mathcal{C}_\textit{base}$, the model can accurately predict labels for nodes within the query set that belong to $\mathcal{C}_{\textit{novel}}$, using only a small number of labeled nodes from the support set.

Inspired by dual-heterogeneity FLHG, researchers in the multi-heterogeneity scenario aim to advance meta-knowledge transfer from a simple one-to-one scenario to a more complex many-to-one scenario. Apart from the challenges in the dual-heterogeneity, a new challenge emerges:

\noindent\textbf{Challenge 5. Mitigating ineffective and negative meta-knowledge transfer (CH5).}
Since base and novel classes originate from multiple HGs with diverse heterogeneities, not all source HGs exhibit structural similarities or relevant heterogeneity for the target HG. Some source HGs may lack meta-knowledge that is useful to the target HG. Learning from these source HGs may result in ineffective or negative knowledge transfer, leading to degraded performance in few-shot learning. Therefore, filtering source HGs to ensure effective knowledge transfer is a critical prerequisite in the multi-heterogeneity scenario.

\section{Research Progress for FLHG}
To correspond with the FLHG problems and challenges mentioned in Section \ref{sec2}, in this section, we comprehensively review current FLHG methods according to their target scenarios, target challenges, and technical perspectives. A detailed categorization of FLHG methods is illustrated in Figure \ref{cate}. In addition, we provide a summary of popular datasets commonly used in FLHG studies.

\begin{figure}
\centering
\scalebox{0.43}{\includegraphics{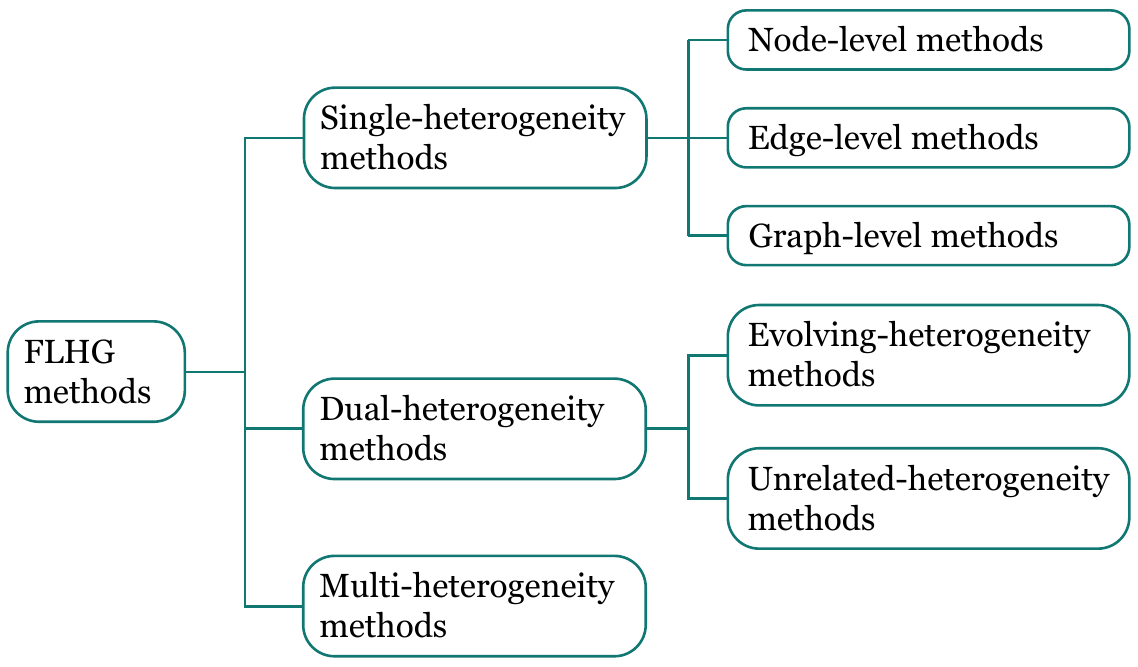}}
\caption{The categorization of FLHG methods.}
\label{cate}
\end{figure}

\subsection{Single-heterogeneity FLHG}
Most FLHG methods focus on the single-heterogeneity scenario, mainly because the simplification of problem modeling and theoretical analysis. Targeting \textbf{CH1} and \textbf{CH2}, these methods have developed several HGNN based few-shot learning frameworks, which extract relevant meta-knowledge for various tasks and improve performance by utilizing the heterogeneous information from unlabeled data. Depending on the specific graph mining task within HGs, single-heterogeneity FLHG methods can be divided into three categories: node-level, edge-level, and graph-level methods.

\noindent\textbf{Node-level Methods.} The node is the fundamental unit from which HGs are constructed. Node-level learning not only facilitates a variety of node-based applications (e.g., node classification and anomaly detection), but also lays the groundwork for advancing to edge-level and graph-level applications. Most node-level methods are designed based on two different few-shot learning frameworks as follows:

\begin{itemize} [leftmargin=*]
    \item \textbf{Gradient-based node-level methods:} These methods typically first employ an HGNN to learn node embeddings. Then, they utilize Model-Agnostic Meta-Learning (MAML) \cite{finn2017model} to optimize the parameters of this HGNN. Finally, they perform node classification using the embeddings obtained from the optimized HGNN. The optimization stage is crucial in these methods. During this stage, tasks from the base classes $\mathcal{C}_\emph{base}$ are used to learn an appropriate initial setting for the HGNN parameters. These parameters are then fine-tuned through several gradient steps over a limited number of labeled nodes from the novel classes $\mathcal{C}_\emph{novel}$. 
    
    Based on the basic paradigm illustrated above, a number of variants have been developed to enhance performance. In particular, HG-Meta \cite{zhang2022hg} improves the optimization stage from a task-oriented view. For intra-task improvements, HG-Meta introduces a task feature scaling module to refine node embeddings. For inter-task improvements, HG-Meta proposes an attention mechanism that leverages meta-path based node degrees to evaluate the importance of tasks, which enables the adjustment of gradient steps according to the importance of each task. In addition, considering that $\mathcal{C}_\emph{base}$ and $\mathcal{C}_\emph{novel}$ may originate from different fields with the same heterogeneity, CrossHG-Meta \cite{zhang2022fewcross} proposes a cross-domain meta-learning approach, which includes a domain critic to align tasks from distinct domains to improve cross-domain generalizability. It also incorporates auxiliary self-supervised learning tasks that deploy positive and negative node pairs to facilitate contrastive regularization.

    \item\textbf{Metric-based node-level methods:} These methods typically adopt the principle of the Prototypical Network (ProNet) \cite{snell2017prototypical}, a simple yet effective framework for few-shot learning. In metric-based node-level methods, an HGNN encoder is first employed to aggregate heterogeneous information into node embeddings. Then, the prototype for each node class is generated by calculating the mean embedding of nodes within the support set. Finally, nodes in the query set are classified by comparing the distance between their embeddings to the prototypes of different classes. Based on this basic model, several enhanced variants have been developed to incorporate heterogeneous information into the prototype calculation. For instance, HPN \cite{hao2023heterogeneous} introduces a semantic attention matrix that quantifies the importance of nodes from different types. This attention matrix is used to compute the weighted summation of support nodes’ embeddings, aiming to form a more refined prototype for each class. 
\end{itemize}

In addition to the work mentioned above, there are several other studies focusing on node-level problems within the single-heterogeneity scenario. For instance, unlike most studies that assume both $\mathcal{C}_\emph{base}$ and $\mathcal{C}_\emph{novel}$ are classes of nodes of the same type, HINFShot \cite{zhuang2021hinfshot} explores scenarios where $\mathcal{C}_\emph{base}$ and $\mathcal{C}_\emph{novel}$ are classes of nodes of different types. HINFShot integrates popular graph neural network (GNN) models (e.g., GAT \cite{velivckovic2017graph}, HGT \cite{hu2020heterogeneous}) with well-known FSL frameworks (e.g., MAML \cite{finn2017model}, ProNet \cite{snell2017prototypical}). Another noteworthy study is MetaHIN \cite{lu2020meta}, which exploits meta-learning on HGs for cold-start recommendation. It proposes a semantic-enhanced task constructor and a co-adaptation meta-learner to learn finer-grained semantic priors for new tasks in both semantic and task-wise manners.

\begin{table*}[]
\caption{A list of representative FLHG methods.}
\label{classification}
\centering
\small
\begin{tabular}{ccccc}
\specialrule{0.15em}{1.75pt}{1.75pt}
\multicolumn{1}{c|}{\textbf{Method}} & \multicolumn{1}{c|}{\textbf{FSL Design}}  & \multicolumn{1}{c|}{\textbf{HGRL Design}}                & \multicolumn{1}{c|}{\textbf{Learning Task}}        & \textbf{Venue} \\\specialrule{0.05em}{1.75pt}{1.75pt}
\multicolumn{5}{c}{\cellcolor[HTML]{E5E5E5}{Single-heterogeneity FLHG}}                                                                                                                                                                     \\\specialrule{0.05em}{1.75pt}{1.75pt}
\multicolumn{1}{c|}{HG-Meta$^{[1]}$}         & \multicolumn{1}{c|}{MAML}                 & \multicolumn{1}{c|}{Meta-path based GNN}                           & \multicolumn{1}{c|}{Node classification}           & SDM'22         \\\specialrule{0.05em}{1.75pt}{1.75pt}
\multicolumn{1}{c|}{CrossHG-Meta$^{[2]}$}    & \multicolumn{1}{c|}{MAML}                 & \multicolumn{1}{c|}{Meta-path based GNN}                           & \multicolumn{1}{c|}{Node classification}           & KDD'22         \\\specialrule{0.05em}{1.75pt}{1.75pt}

\multicolumn{1}{c|}{HPN$^{[3]}$}             & \multicolumn{1}{c|}{ProNet} & \multicolumn{1}{c|}{Node type based GNN}                           & \multicolumn{1}{c|}{Node classification}           & ICONIP'23      \\\specialrule{0.05em}{1.75pt}{1.75pt}

\multicolumn{1}{c|}{HINFShot$^{[4]}$}             & \multicolumn{1}{c|}{ProNet, MAML} & \multicolumn{1}{c|}{GAT, HGT}                           & \multicolumn{1}{c|}{Node classification}           & ICMR'21      \\\specialrule{0.05em}{1.75pt}{1.75pt}

\multicolumn{1}{c|}{MetaHIN$^{[5]}$}         & \multicolumn{1}{c|}{MAML}                 & \multicolumn{1}{c|}{Edge type \& meta-path based GNN}              & \multicolumn{1}{c|}{User-item recommendation}      & KDD'20         \\\specialrule{0.05em}{1.75pt}{1.75pt}
\multicolumn{1}{c|}{GMatching$^{[6]}$}       & \multicolumn{1}{c|}{MatchNet}             & \multicolumn{1}{c|}{Neighbor encoder}                    & \multicolumn{1}{c|}{Relation prediction}           & EMNLP’18       \\\specialrule{0.05em}{1.75pt}{1.75pt}
\multicolumn{1}{c|}{FSRL$^{[7]}$}            & \multicolumn{1}{c|}{MatchNet}             & \multicolumn{1}{c|}{Neighbor encoder}                    & \multicolumn{1}{c|}{Relation prediction}           & AAAI’20        \\\specialrule{0.05em}{1.75pt}{1.75pt}
\multicolumn{1}{c|}{FIRE$^{[8]}$}            & \multicolumn{1}{c|}{MAML}                 & \multicolumn{1}{c|}{Neighbor encoder}                    & \multicolumn{1}{c|}{Multi-hop relation prediction} & EMNLP’20       \\\specialrule{0.05em}{1.75pt}{1.75pt}
\multicolumn{1}{c|}{Meta-MGNN$^{[9]}$}       & \multicolumn{1}{c|}{MAML}                 & \multicolumn{1}{c|}{GIN based molecular encoder}         & \multicolumn{1}{c|}{Molecule classification}       & WWW’21         \\\specialrule{0.05em}{1.75pt}{1.75pt}
\multicolumn{1}{c|}{MTA$^{[10]}$}             & \multicolumn{1}{c|}{MAML}                 & \multicolumn{1}{c|}{Motif encoder}                               & \multicolumn{1}{c|}{Molecule classification}       & SDM'23         \\\specialrule{0.05em}{1.75pt}{1.75pt}
\multicolumn{5}{c}{\cellcolor[HTML]{E5E5E5}{Dual-heterogeneity FLHG}}                                                                                                                                                                       \\\specialrule{0.05em}{1.75pt}{1.75pt}
\multicolumn{1}{c|}{COHF$^{[11]}$}            & \multicolumn{1}{c|}{ProNet} & \multicolumn{1}{c|}{Variational autoencoder}             & \multicolumn{1}{c|}{Node classification}           & -              \\\specialrule{0.05em}{1.75pt}{1.75pt}
\multicolumn{1}{c|}{CGFL$^{[12]}$}            & \multicolumn{1}{c|}{ProNet} & \multicolumn{1}{c|}{Affiliation \& interaction relation encoder} & \multicolumn{1}{c|}{Node classification}           & CIKM'23        \\\specialrule{0.05em}{1.75pt}{1.75pt}
\multicolumn{5}{c}{\cellcolor[HTML]{E5E5E5}{Multi-heterogeneity FLHG}}                                                                                                                                                                   \\\specialrule{0.05em}{1.75pt}{1.75pt}
\multicolumn{1}{c|}{MetaGS$^{[13]}$}          & \multicolumn{1}{c|}{ProNet} & \multicolumn{1}{c|}{Semantic subgraph based GNN}                   & \multicolumn{1}{c|}{Semantic relation prediction}  & TKDE'23  \\\specialrule{0.15em}{1.75pt}{1.75pt}
\end{tabular}
\\
\raggedright{Note: $^{[1]}$\cite{zhang2022hg}; $^{[2]}$\cite{zhang2022fewcross}; $^{[3]}$\cite{hao2023heterogeneous}; 
$^{[4]}$\cite{zhuang2021hinfshot}; $^{[5]}$\cite{lu2020meta}; $^{[6]}$\cite{xiong2018one}; $^{[7]}$\cite{zhang2020few}; $^{[8]}$\cite{zhang2020fewCC}; $^{[9]}$\cite{guo2021few}; $^{[10]}$\cite{meng2023meta}; $^{[11]}$\cite{ding2024few}; $^{[12]}$\cite{ding2023cross}; $^{[13]}$\cite{ding2023few};
}
\end{table*} 

\noindent\textbf{Edge-level Methods.}
Generally, edge-level methods first adopt an HGNN-based relation encoder to learn representations of relations by aggregating the embeddings of corresponding node pairs. This encoder is then integrated into a few-shot learning framework to facilitate the prediction of relation classes using few-labeled samples. Specifically, on the one hand, some methods are based on the Matching Network (MatchNet) framework \cite{vinyals2016matching}, which calculates a matching score between the relation embeddings and query samples. This score is utilized to evaluate the acceptability of each query sample. In particular, GMatching \cite{xiong2018one} is the first to employ MatchNet for one-shot relation prediction in knowledge graphs. Building on this, FSRL \cite{zhang2020few} enhances GMatching by attentively aggregating features of support samples for each relation and improving node embeddings through a heterogeneous neighbor aggregator. Moreover, REFORM \cite{wang2021reform} introduces an error mitigation module to reduce errors in the construction of knowledge graphs.

On the other hand, some methods adopt the MAML framework \cite{finn2017model} to optimize parameters in the relation encoder. Specifically, MetaKGR \cite{lv2019adapting} adopts MAML to learn effective parameters from high-frequency relations. FIRE \cite{zhang2020fewCC} extends Meta-KGR with by incorporating a heterogeneous neighbor aggregator and a search space pruning strategy. Meta-iKG \cite{zheng2022subgraph} translates link prediction as a subgraph modeling problem and utilizes local subgraphs to transfer subgraph-specific information and to rapidly learn transferable patterns.


\noindent\textbf{Graph-level Methods.}
Graph-level methods typically first adopt an HGNN with graph pooling techniques \cite{liu2022graph} to learn representations at the graph level, and then combine the HGNN with various few-shot learning frameworks to classify graph/subgraphs using a limited number of labeled HGs. Existing methods primarily focus on the molecular property prediction problem (i.e., molecular graph classification). For instance, Meta-MGNN \cite{guo2021few} utilizes graph-level HGNNs to learn the embeddings of individual molecules and introduces task weights via MAML for more effective optimization. PAR \cite{wang2021property} enhances Meta-MGNN by capturing the relational structure among different molecular properties, which facilitates effective label propagation among similar molecules. Recently, to tackle the issues of memorization and learner overfitting within the few-shot learning framework, MTA \cite{meng2023meta} employs a convex combination of structural motifs, which are common substructures in molecules, to facilitate efficient online task augmentations in each sampled task.

\subsection{Dual-heterogeneity FLHG} 
In the scenario of dual-heterogeneity, base classes and novel classes exist within distinct heterogeneous environments. The variance in heterogeneity renders the conventional FLHG framework, which relies on transferring knowledge through shared node/edge types or meta-paths, infeasible. Consequently, two new challenges arise in this scenario, i.e., \textbf{CH3} and \textbf{CH4}. Targeting these challenges, dual-heterogeneity methods mainly focus on either mitigating distribution shifts between the source HG and the target HG, or exploring the underlying commonalities and connections between the two different heterogeneities. Accordingly, these methods can be divided into two categories: evolving-heterogeneity methods and unrelated-heterogeneity methods.

\begin{table*}[]
\caption{A list of commonly used and publicly accessible real-world datasets for FLHG.}
\centering
\small
\label{datasets}
\begin{tabular}{c|c|c|c|c|c}
\specialrule{0.15em}{1.75pt}{1.75pt}
\textbf{Dataset} & \textbf{Domain} & \textbf{\# Nodes}   & \textbf{Task}        & \textbf{Reference} & \textbf{Link}                             \\ \specialrule{0.05em}{1.75pt}{1.75pt}
AMiner           & Academic        & 164,220                 & Node classification  & \cite{zhang2022fewcross}       & \url{https://shorturl.at/bjKS0}                 \\ \specialrule{0.05em}{1.75pt}{1.75pt}
U.S. Patents     & Business        & 55,641                 & Node classification  & \cite{zhang2022fewcross}       & \url{https://shorturl.at/irtIN}                 \\ \specialrule{0.05em}{1.75pt}{1.75pt}
Amazon           & E-commerce      & 108,022                   & Node classification  & \cite{zhang2022fewcross}       & \url{https://shorturl.at/hEHJ2}                 \\ \specialrule{0.05em}{1.75pt}{1.75pt}
IMDB             & Movie           & 109,573           & Node classification  & \cite{hao2023heterogeneous}                & \url{https://shorturl.at/flL19}                 \\ \specialrule{0.05em}{1.75pt}{1.75pt}
Douban movie     & Movie           & 31,157           & Node classification  & \cite{hao2023heterogeneous}                & \url{https://shorturl.at/efzHL}                 \\ \specialrule{0.05em}{1.75pt}{1.75pt}
Douban book      & Book            & 42,070               & Node classification  & \cite{lu2020meta}            & \url{https://shorturl.at/diz35}                 \\ \specialrule{0.05em}{1.75pt}{1.75pt}
MovieLens        & Movie           & 20,137     & Node classification  & \cite{lu2020meta}            & \url{https://shorturl.at/ktwQW} \\ \specialrule{0.05em}{1.75pt}{1.75pt}
YELP             & Business        & 86,874   & Node classification  & \cite{lu2020meta}            & \url{https://shorturl.at/hiwCR}                 \\ \specialrule{0.05em}{1.75pt}{1.75pt}
NELL             & Knowledge base  & 68,545                         & Relation prediction  & \cite{xiong2018one}          & \url{https://shorturl.at/aerxQ}                 \\ \specialrule{0.05em}{1.75pt}{1.75pt}
Wiki             & Knowledge base  & 4,838,244                         & Relation prediction  & \cite{xiong2018one}          & \url{https://shorturl.at/chiJ8}                 \\ \specialrule{0.05em}{1.75pt}{1.75pt}
Tox21            & Biology         & 7,831                   & Graph classification & \cite{guo2021few}          & \url{https://shorturl.at/asSWX}               \\ \specialrule{0.05em}{1.75pt}{1.75pt}
Sider            & Biology         & 1,427                   & Graph classification & \cite{guo2021few}          & \url{https://shorturl.at/aHNO1}               \\ \specialrule{0.05em}{1.75pt}{1.75pt}
YELP restaurant  & Business        & 3,913 & Node classification  & \cite{ding2024few}               & \url{https://shorturl.at/efzHL}                 \\ \specialrule{0.05em}{1.75pt}{1.75pt}
PubMed           & Medical         & 63,109  & Node classification  & \cite{ding2023cross}               & \url{https://shorturl.at/nwLSX}                 \\ \specialrule{0.05em}{1.75pt}{1.75pt}
Facebook         & Social network  & 5,307  & Relation prediction  & \cite{ding2023few}             & \url{https://shorturl.at/apruI}     \\\specialrule{0.15em}{1.75pt}{1.75pt}    
\end{tabular}
\end{table*}

\noindent\textbf{Evolving-heterogeneity Methods:} 
These methods target the scenarios where the target HG $\emph{G}_\emph{tgt}$ evolves from the source HG $\emph{G}_\emph{src}$. For instance, $\emph{G}_\emph{src}$ might be an initial social network with “user” nodes and “friendship” edges. Over time, $\emph{G}_\emph{tgt}$ might expand to include new node types like “organization” and new edge types such as “affiliation”, while the initial “friendship” edges may diminish. Typically, these methods investigate the evolutionary process and the distribution shifts between the two HGs to identify invariant features that facilitate meta-knowledge transfer.

Recently, \cite{ding2024few} introduces the COHF model for node classification. Specifically, COHF utilizes distribution shifts to model the evolution from $\emph{G}_\emph{src}$ to $\emph{G}_\emph{tgt}$, conceptualizing the dual-heterogeneity problem as an out-of-distribution generalization problem. Utilizing principles from causal learning \cite{pearl2009causality}, COHF introduces a structural causal model to characterize distribution shifts in HGs. Then, it employs a variational autoencoder-based HGNN to mitigate the impact of these shifts and combines the HGNN with ProNet to transfer invariant knowledge between the two HGs.

\noindent\textbf{Unrelated-heterogeneity Methods:} 
These methods target the scenarios where $\emph{G}_\emph{src}$ and $\emph{G}_\emph{tgt}$ originate from different systems and have completely distinct types of nodes and edges. In such cases, it is crucial to identify and transfer generalized knowledge that is independent of specific node and edge types. Therefore, existing methods typically transform specific heterogeneous information into more abstract features. This transformation aids in uncovering commonalities among these abstract features, thus facilitating the identification and extraction of suitable meta-knowledge. For instance, CGFL \cite{ding2023cross} leverages affiliation and interaction relations, which are prevalent across HGs with varying heterogeneities. By utilizing an HGNN to learn various meta-patterns emerging from these relations, CGFL is able to derive meta-knowledge applicable to few-shot learning.

\subsection{Multi-heterogeneity FLHG}
Although multi-heterogeneity FLHG is inspired by dual-heterogeneity FLHG, it aims to explore more general meta-knowledge that can be widely applied across diverse heterogeneities. Apart from the challenges (\textbf{CH3} and \textbf{CH4}) present in the dual-heterogeneity scenario, multi-heterogeneity scenario introduces a new challenge, i.e., ineffective and negative meta-knowledge transfer (\textbf{CH5}).

Multi-heterogeneity FLHG is a challenging scenario with few solutions \cite{ding2023cross,ding2023few} proposed to date. In \cite{ding2023cross}, the authors propose a three-level score module to evaluate the transferability of source HGs, the consistency of few-shot tasks, and the informativeness of labeled samples. They also introduce a multi-view approach that utilizes multiple heterogeneity-independent perspectives to extract comprehensive and generalized knowledge across HGs. Meanwhile, the MetaGS approach proposed in \cite{ding2023few} focuses on the semantic relation prediction problem. MetaGS decomposes the complex semantics inherent in specific heterogeneities into common and fundamental semantics. It then transfers knowledge about these basic semantics to facilitate the few-shot learning across HGs with various heterogeneities.

\subsection{Summary of Datasets}
Datasets are essential for the empirical analysis of FLHG approaches. To support the empirical evaluation of the methods reviewed, in Table \ref{datasets} we list public and real-world datasets with different characteristics from various domains. These datasets are widely used in assessing FLHG methods and are publicly available. We also provide references to notable studies that have used these datasets for FLHG studies.

\section{Future Directions}
Few-shot learning on heterogeneous graphs is an emerging research topic. Although significant progresses have been made for FLHG, there still remain plenty of research directions worthy of future explorations.

\subsection{Open Problems in Three Scenarios}
In the three main scenarios of FLHG, each scenario still contains valuable problems that remain unexplored.

\begin{itemize}[leftmargin=*]
\item In the single-heterogeneity scenario, existing studies primarily focus on node-level and edge-level problems, overlooking graph-level problems across various domains. For instance, detecting communities in social networks, analysing purchase trends in e-commerce networks, and predicting traffic congestion in transportation networks. These problems are of significant practical importance and need further research.

\item In the dual-heterogeneity scenario, current research mainly targets problems at the node-level, leaving edge-level and graph-level problems have not been thoroughly investigated. A promising direction within this scenario is the development of general frameworks to tackle problems at the edge or graph levels. Such frameworks can serve as guidance for further specific applications, such as relation prediction in knowledge graphs and molecular classification in biology networks.

\item In the multi-heterogeneity scenario, existing methods tend to simplify the complex many-to-one scenario into multiple simpler one-to-one scenarios for analysis, which demands significant computational resources. Therefore, it is essential to develop novel frameworks that can directly and efficiently deal with the many-to-one scenario.
\end{itemize}

\subsection{Dynamics of Heterogeneous Graphs}
Most FLHG methods focus on static HGs \cite{zhang2022fewcross,ding2023cross}. However, in the real world, networks often change over time, with frequent additions and deletions of nodes and edges. This dynamic nature of networks makes most existing FLHG methods less effective, because they struggle to adapt to such changes. Therefore, there is a research direction for developing a novel few-shot learning framework, which can extract meta-knowledge about both structural and temporal dependencies across various dynamic HGs.

\subsection{Multi-aspect Auxiliary Tasks}
Due to the rich semantic information contained within HGs, combining few-shot learning with auxiliary tasks that use unlabeled data can improve the performance of FLHG methods. However, most FLHG methods do not utilize auxiliary tasks and are limited to specific task spaces within a predefined model structure. For instance, in node-classification FLHG methods, only tasks directly related to node classification are considered. Although some FLHG methods have started to incorporate auxiliary self-supervised learning tasks \cite{zhang2022hg,zhang2022fewcross}, they typically focus on tasks from a single aspect of HGs, such as self-supervised link prediction at the edge-level. This highlights the potential for a novel framework that can tackle various auxiliary tasks at the node-level, edge-level, and graph-level simultaneously. Moreover, exploring effective strategies for optimizing primary few-shot tasks with these auxiliary tasks is a potential area for future studies.

\subsection{Explainability and Robustness} 
In the real world, many FLHG applications are risk-sensitive, such as few-shot fraud detection \cite{ding2021few} and low-data drug discovery \cite{lv2023meta}. These applications often require model explainability to improve end-user trust and to guide model design by identifying which parts of FLHG significantly impact performance. Recently, several studies have explored the explainability of graph learning models. For instance, graph causal learning \cite{sui2022causal} combines graph theory with causal inference to model cause-and-effect relationships in graphs. Meanwhile, disentangled learning \cite{wang2020disenhan,hou2021disentangled} divides graph representation into distinct latent spaces to improve explainability. However, generalizing these techniques to FLHG methods is non-trivial and deserves further research.

In addition, model robustness is crucial for various FLHG applications. On the one hand, since real-world HG data is often noisy and inconsistent, robust models are essential to ensure reliable performance, especially when the few-labeled data is unrepresentative or contains outliers \cite{xu2021robustness}. On the other hand, as HG-based models are increasingly deployed in sensitive and high-stakes environments, they are vulnerable to adversarial attacks that seek to alter model outcomes \cite{zhang2020gnnguard}. However, most FLHG models rely on high-quality HG data with consistent distributions and no adversarial intentions. Therefore, developing robust FLHG models that can handle inconsistent and high-risk HG data is a valuable direction for future exploration.

\subsection{Utilizing Large Language Models}
Recently, the increasing capabilities of Large Language Models (LLMs) in understanding and utilizing human knowledge have demonstrated significant potential across various fields \cite{kaddour2023challenges}. Since the generation of most HGs heavily relies on real-world knowledge, applying LLMs to FLHG has become an increasingly attractive prospect. On the one hand, LLMs can provide valuable prior knowledge to enhance feature extraction from a limited number of samples. For instance, in social networks, LLMs can perform comprehensive analysis of user profiles to identify unique characteristics. These characteristics can then be used to generate higher-quality node features for the FLHG model. On the other hand, LLMs are capable of utilizing related knowledge to evaluate the importance of each few-labeled sample, thereby enabling more accurate learning of class features. Consequently, integrating LLMs with FLHG methods represents a promising avenue for future research.



\section{Conclusions}
This paper presents the first review of the advancements in the emerging research area of FLHG. We first summarize existing methods into three categories based on three types of heterogeneity environments, i.e., single-heterogeneity, dual-heterogeneity, and multi-heterogeneity. Then, we propose a comprehensive analysis of the existing studies within each category. Finally, we identify and discuss several promising directions for future FLHG research. We hope this review will serve as a useful reference for researchers in FLHG.


\bibliographystyle{named}
\bibliography{ijcai24}

\end{document}